\documentclass{article}
\usepackage{spconf,amsmath,graphicx,xcolor,cite,amssymb}
\usepackage{enumitem,kantlipsum}

\def\etal{\emph{et al.}}

\makeatletter
\def\old@comma{,}
\catcode`\,=13
\def,{%
  \ifmmode%
    \old@comma\discretionary{}{}{}%
  \else%
    \old@comma%
  \fi%
}
\makeatother

\title{Audio-Visual Deepfake Detection With Local Temporal Inconsistencies}

\name{Marcella Astrid$^\star$ \qquad
Enjie Ghorbel$^{\star \dagger}$ \qquad Djamila Aouada$^\star$ \thanks{This work was supported by the Luxembourg National Research Fund (FNR) under the project BRIDGES2021/IS/16353350/FaKeDeTeR and POST Luxembourg.}}
\address{$^\star$Interdisciplinary Centre for Security, Reliability and Trust (SnT), \\  University of Luxembourg, Luxembourg \\
$^\dagger$Cristal Laboratory, National School of Computer Sciences (ENSI), \\ Manouba University, Tunisia \\}
\begin{document}

\setlength{\abovedisplayskip}{8pt}
\setlength{\belowdisplayskip}{8pt}
%\ninept
%
\maketitle
\begin{abstract}
This paper proposes an audio-visual deepfake detection approach that aims to capture fine-grained temporal inconsistencies between audio and visual modalities. To achieve this, both architectural and data synthesis strategies are introduced. From an architectural perspective, a temporal distance map, coupled with an attention mechanism, is designed to capture these inconsistencies while minimizing the impact of irrelevant temporal subsequences. Moreover, we explore novel pseudo-fake generation techniques to synthesize local inconsistencies. Our approach is evaluated against state-of-the-art methods using the DFDC and FakeAVCeleb datasets, demonstrating its effectiveness in detecting audio-visual deepfakes.
\end{abstract}

\vspace{-1mm}
\section{Introduction}
\label{sec:intro}
\vspace{-2mm}

% Audio-visual deepfake video generation can be beneficial but poses risks when used maliciously \cite{Scam_news,Disinformation_news}. As deepfake quality improves, distinguishing real from fake becomes harder.  Hence, it is crucial to develop a system for detecting audio-visual deepfakes.
Audio-visual deepfakes have beneficial applications but pose significant risks when misused~\cite{Scam_news,Disinformation_news}. As their quality improves, distinguishing real from fake becomes increasingly difficult, highlighting the need for effective detection systems.

One approach for detecting audio-visual deepfakes is to identify inconsistencies between audio and visual data, as shown in numerous research works~\cite{chugh2020not,gu2021deepfake,feng2023self,astrid2024statistics}. However, since deepfake artifacts are mostly subtle~\cite{nguyen2024laa,zhao2021multi}, a fine-grained classification is needed. The use of a fine-grained approach in audio-visual deepfake detection has only recently been investigated in~\cite{astrid2024detecting} by exploring both architecture design and data augmentation strategies. However, Astrid \etal~\cite{astrid2024detecting} focus only on spatial inconsistencies, while exploring a very limited range of data augmentation techniques. To the best of our knowledge, no previous work has explored temporal fine-grained artifacts despite their informativeness in audio-visual deepfake detection. It can be noted that this line of work is distinct from fusion-based methods (e.g.,~\cite{yang2023avoid,wang2024avt2,ilyas2023avfakenet}) that aim to fuse multimodal information, as well as from methods that extract the identity of individuals (e.g.,~\cite{cheng2023voice,cai2022you}).

\begin{figure*}[tb]
  \centering
  \includegraphics[width=0.9\linewidth]{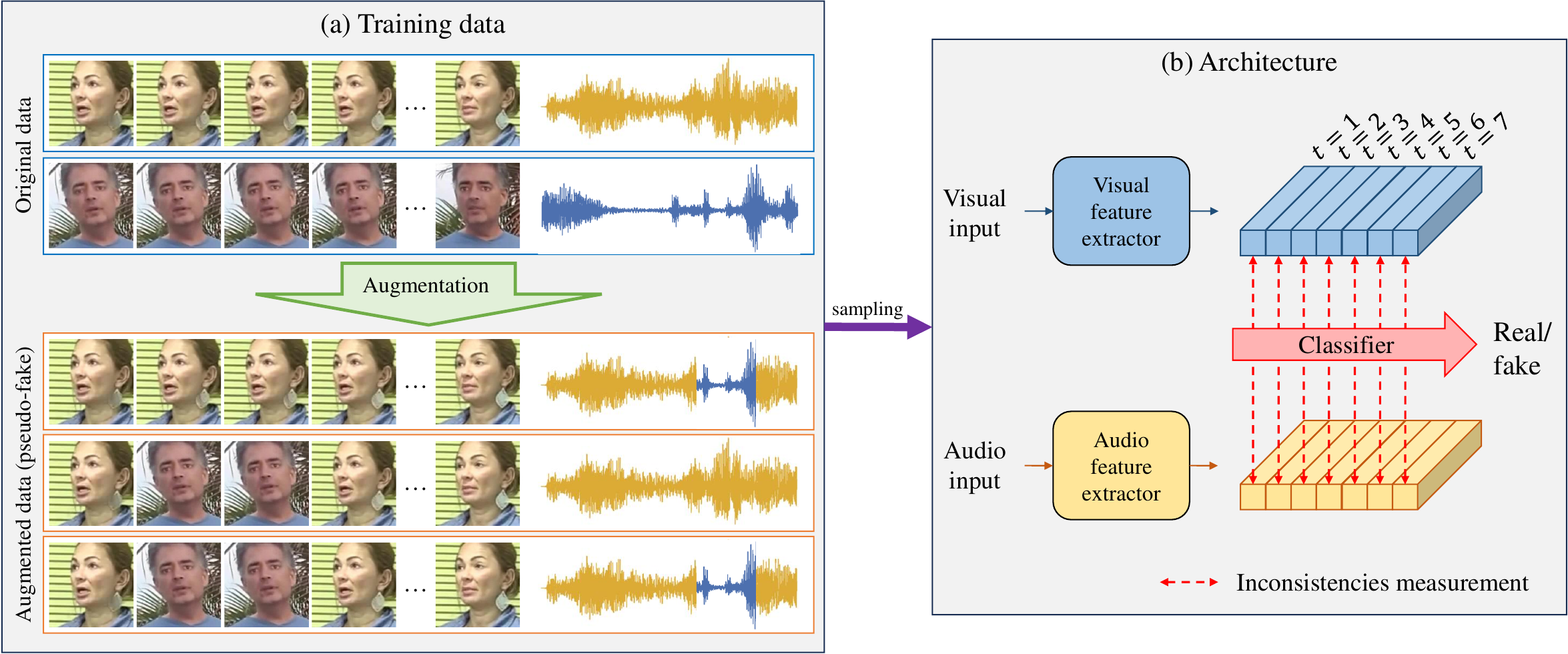}
  \vspace{-3mm}
  \caption{We address temporal fine-grained inconsistencies from (a) the data synthesis perspective and (b) the architectural perspective. (a) We augment the original data with pseudo-fake examples by locally modifying the data in the temporal domain. (b) The classifier evaluates audio-visual inconsistencies at each temporal step.}
  \vspace{-4mm}
  \label{fig:introduction}
\end{figure*}

In this paper, we propose therefore to introduce mechanisms for modeling temporal fine-grained irregularities from both the architecture and the data synthesis perspectives.
%Additionally, for data augmentation, we explore a wider variety of pseudo-fake data synthesis methods compared to~\cite{astrid2024detecting}.
Specifically, in terms of architecture, we compute temporal distances for each time step between audio and visual data. These fine-grained temporal distances are fed into the classifier, as shown in Fig.~\ref{fig:introduction}(b). Our architecture also incorporates an attention mechanism to minimize the impact of irrelevant audio-visual distances, such as background sounds.
% To be more specific, in terms of architecture, the classifier in \cite{astrid2024detecting} uses the distance between audio and local spatial patches to detect deepfakes. We believe this concept can be extended to the temporal domain. Thus, we compute local temporal distances between audio and visual data and input them into the classifier, as shown in Fig.~\ref{fig:introduction}(b). Our architecture also incorporates an attention mechanism to minimize the impact of irrelevant audio-visual distances, such as background sounds.
Moreover, we propose extending the work of \cite{astrid2024detecting} where additional fake data are generated by replacing several frames from a given audio/video with data coming from a different audio/video, as shown in Fig.~\ref{fig:introduction}(a). In this work, to synthesise pseudo-fakes with subtle temporal artifacts, we suggest  replacing sub-sequences from an original video with slightly manipulated versions of the same video using simple operations such as translation, flip, and frame repetition.

In summary, our contributions are as follows:
1) To the best of our knowledge, we are among the first to explore temporal local inconsistencies for detecting audio-visual deepfakes using both data augmentation and architecture strategies; 
2) We propose augmenting the training data by generating pseudo-fake examples with subtle inconsistencies and exploring different methods for their generation;
3) We design a deepfake classifier that measures audio-visual feature distances across time to capture fine-grained temporal inconsistencies, enhanced with an attention mechanism; 
4) We evaluate our method against state-of-the-art (SOTA) approaches using the DFDC dataset~\cite{dolhansky2020deepfake} for in-dataset and the FakeAVCeleb dataset~\cite{khalid2021fakeavceleb} for cross-dataset settings.

\noindent \textbf{Paper organization.} Our methodology is detailed in Section~\ref{sec:method}, while the experiments and the results are detailed in Section~\ref{sec:experiments}.  Finally, the conclusion is given in Section~\ref{sec:conclusion}.

\vspace{-3mm}
\section{Methodology}
\label{sec:method}
\vspace{-2mm}

As shown in Fig.~\ref{fig:introduction}, we address the fine-grained deepfake detection problem at the temporal level from two perspectives, namely: (1) data augmentation with pseudo-fake generation (Section~\ref{subsec:augmentation}); and (2) architectural design for capturing fine-grained audio-visual distances in the temporal dimension (Section~\ref{subsec:architecture}). 

\subsection{Data augmentation}
\label{subsec:augmentation}

The original training set consists of pairs of audio-visual data, $\mathbf{X}^v$ and $\mathbf{X}^a$, with sizes $T^v \times C^v \times H^v \times W^v$ and $T^a \times C^a$, respectively. Here, the superscripts $v$ and $a$ denote visual and audio data, respectively. $T$, $C$, $H$, and $W$ represent time, channel, height, and width dimensions, respectively. Since we use a waveform input for the audio, $C^a$ equals $1$. The audio and video sequences are extracted from the same part of the overall video.

We generate pseudo-fakes from visual $\mathbf{P}^v$ and audio $\mathbf{P}^a$ data to augment the original training set. Each pseudo-fake data point is randomly selected as either $\{\mathbf{X}^v, \mathbf{P}^a\}$, $\{\mathbf{P}^v, \mathbf{X}^a\}$, or $\{\mathbf{P}^v, \mathbf{P}^a\}$, and is labeled as \textit{fake}. We use pseudo-fake data as input instead of the original data with a probability of $0.5$.

We generate $\mathbf{P}^v$ and $\mathbf{P}^a$ in a similar way. For the sake of simplicity, we use a shared variable $\mathbf{A} = \{\mathbf A_1, \mathbf A_2, \ldots, \mathbf A_T\}$ to represent both $\mathbf{X}^v$ and $\mathbf{X}^a$, with $T=T^v$ for visual data and $T=T^a$ for audio data. Each $\mathbf A_i$ represents a frame of size $C^v \times H^v \times W^v$ for visual data or a waveform magnitude value for audio data. The input $\mathbf{A}$ is then manipulated to create the pseudo-fake.

\begin{figure*}[tb]
  \centering
  \includegraphics[width=0.9\linewidth]{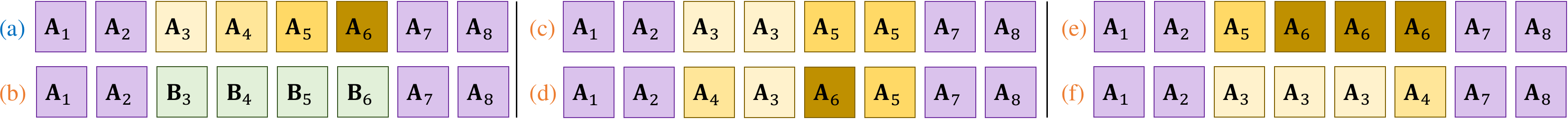}
  \vspace{-2mm}
  \caption{Given (a) an original data sequence, we generate pseudo-fake data by locally modifying the sequence. In this example, we modify $\mathbf A_3$ to $\mathbf A_6$. We explore various modifications: (b) replacing with sub-sequences from another data sample, (c) repeating, (d) flipping, and (e-f) translating from the left or right.}
  \vspace{-4mm}
  \label{fig:method_pseudofake}
\end{figure*}

\vspace{-2mm}
\subsubsection{Temporally-local manipulation}

Following~\cite{astrid2024detecting}, to create a pseudo-fake $\mathbf{P}$ with subtle artifacts from $\mathbf{A}$, we locally modify the sequence from the time step $i$ to $i+l-1$. We denote this modified chunk as $\mathbf{C} = \mathbf A_{i:i+l-1} = \{\mathbf A_i, \mathbf A_{i+1}, \ldots, \mathbf A_{i+l-1}\}$. For example, in Fig.~\ref{fig:method_pseudofake}(a), $i$ and $l$ are 3 and 4, respectively, so $\mathbf{C} = \{\mathbf A_3, \mathbf A_4, \mathbf A_5, \mathbf A_6\}$. The modification length $l$ is randomly selected within $[l_{min}, l_{max}]$, where
\begin{equation}
    l_{min} = r_{min} \times T\text{,} \qquad l_{max} = r_{max} \times T \text{,}
\end{equation}
with $r_{min}$ and $r_{max}$ being hyperparameters in the range $]0, 1]$. We set the minimum $l_{min}$ to 2.

\subsubsection{Manipulation types}
\label{subsubsec:manipulation}

There can be several ways to modify $\mathbf{C}$ so that the resulting sequence differs from the original $\mathbf{A}$:
\begin{enumerate}[leftmargin=*]
    \item Replacing with another clip
    % \begin{equation}
    %     \mathbf C = \mathbf B_{i+a}  \quad \text{where } a = \{0, 1, ..., l-1 \}   \text{,}
    % \end{equation}
    \begin{equation}
        \mathbf C = \mathbf B_{i:i+l-1}   \text{,}
    \end{equation}
    where $\mathbf B_{i:i+l-1}$ is a subsequence from a randomly selected data sample. This method, depicted in Fig.~\ref{fig:method_pseudofake}(b) is used in~\cite{astrid2024detecting}. In this work, we also  consider other methods.
    % For example, if $l=7$, $P=\{B_{i}, B_{i+1}, B_{i+2}, B_{i+3}, B_{i+4}, B_{i+5}, B_{i+6}\}$

    \item Repeating
    % \begin{equation}
    %     P = A_{i+\left \lfloor{\frac{a}{r}}\right \rfloor * r} \quad \text{where } a = \{0, 1, ..., \left \lfloor{\frac{l}{r}}\right \rfloor * r + l\bmod r-1 \}
    % \end{equation}
    \begin{equation}
        \mathbf C = \mathbf A_{i+\left \lfloor{\frac{a}{p}}\right \rfloor * p} \quad \text{where } a = \{0, 1, ..., l-1 \} \text{,}
    \end{equation}
    where $p$ is the number of repetitions, randomly chosen from $2$ to $l$ for each generated pseudo-fake. In the example of Fig.~\ref{fig:method_pseudofake}(c),  we use $p=2$.
    
    % For example, if repeat $r=3$ and pseudo fake length $l=7$, $a = \{0, 1, ..., 6\}$, so \\ $P = \{A_{i}, A_{i}, A_{i}, A_{i+3}, A_{i+3}, A_{i+3}, A_{i+6}\}$.

    \item Flipping
    \begin{equation}
        \mathbf C =  \mathbf A_{i+ 2f \left \lfloor{\frac{a}{f}}\right \rfloor + f - 1 - a} \quad \text{where } a = \{0, 1, ..., l-1 \} \text{,}
    \end{equation}
    where $f$ is the flipping frequency, randomly chosen from $2$ to $l$ for each pseudo-fake. In the example of Fig.~\ref{fig:method_pseudofake}(d), we uses $f=2$.
    % For example, if flip $f=3$ and pseudo fake length $l=7$, $a = \{0, 1, ..., 6\}$, so \\ $P = \{A_{i+2}, A_{i+1}, A_{i}, A_{i+5}, A_{i+4}, A_{i+3}, A_{i+8}\}$.

    % \item Random
    % \begin{equation}
    %     P = \text{random}(A_i, A_{i+1}, ..., A_{i+l-2}, A_{i+l-1})
    % \end{equation}
    % For example, $P = \{A_{i+2}, A_{i}, A_{i+6}, A_{i+5}, A_{i+3}, A_{i+1}, A_{i+4}\}$.

    \item Translating left or right
    \begin{equation}
        \mathbf C = \mathbf A_{i + \min(l-1, a+v)}  \text{ where } a = \{0, 1, ..., l-1 \} \text{, or}
    \end{equation}
    \begin{equation}
        \mathbf C = \mathbf A_{i + \max(0, a-v)}  \quad \text{where } a = \{0, 1, ..., l-1 \}   \text{,}
    \end{equation}
    % For example, if $v=3$ and $l=7$, $P=\{A_i, A_i, A_i, A_{i+1}, A_{i+2}, A_{i+3}, A_{i+4}\}$
    where $v$ is the translation step, randomly selected from $2$ to $l$. The direction, left or right, is also chosen randomly. Examples in Fig.~\ref{fig:method_pseudofake}(e) and (f) use $v=2$.
    % We use in the examples of Fig.~\ref{fig:method_pseudofake}(e) and (f) $v=2$.
    % For example, if $v=3$ and $l=7$, $P=\{A_{i+3}, A_{i+4}, A_{i+5}, A_{i+6}, A_{i+6}, A_{i+6}, A_{i+6}\}$

\end{enumerate}

\vspace{-2mm}
\subsection{Architectural design}
\label{subsec:architecture}

\begin{figure*}[tb]
  \centering
  \includegraphics[width=0.9\linewidth]{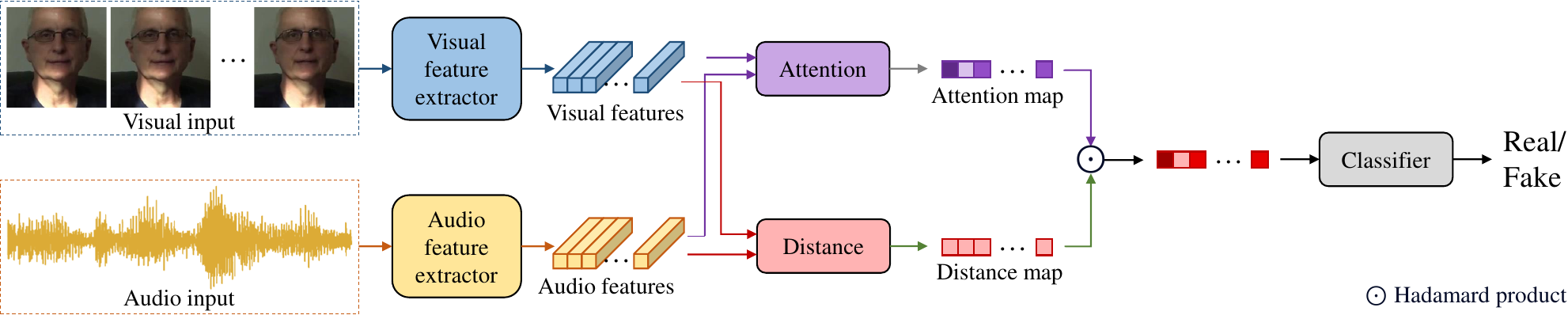}
  \vspace{-3mm}
  \caption{Our architecture uses a fine-grained distance map for each time step of the extracted features, combined with an attention mechanism, to classify whether the input pair is fake.}
  \vspace{-2mm}
  \label{fig:method_architecture}
\end{figure*}

To capture subtle temporal inconsistencies, we design a classifier that processes the distance map at each time step. The architecture, shown in Fig.~\ref{fig:method_architecture}, includes feature extractors, distance calculation, an attention mechanism, and a classifier.

\vspace{-3mm}
\subsubsection{Feature extractors}

Given a pair of audio-visual inputs $\mathbf{X}^v$ (or $\mathbf{P}^v$) and $\mathbf{X}^a$ (or $\mathbf{P}^a$), we extract features as follows:
\begin{equation}
    \mathbf{F}^v = \mathcal F^v(\mathbf{X}^v)\text{,}  \qquad \mathbf{F}^a = \mathcal F^a(\mathbf{X}^a) \text{,}
\end{equation}
where $\mathbf{F}^v$ and $\mathbf{F}^a$ are features of size $T' \times C'$, with $T'$ and $C'$ representing the time and channel dimensions. The visual feature extractor $\mathcal{F}^v(\cdot)$ is a ResNet-based 3D Convolution (Conv3D) model, designed to be shallow to preserve a larger temporal dimension. We then use a 3D adaptive average pooling block to output features with a spatial size of 1 and a temporal dimension equal to $T'$. The audio feature extractor $\mathcal{F}^a(\cdot)$ is based on 1D Convolution (Conv1D) layers to produce features of size $T' \times C'$.

\vspace{-2mm}
\subsubsection{Temporal local distance}
To identify fine-grained temporal inconsistencies, we compute a distance map $\mathbf{m}$ of size $T'$, where each element represents the L2 distance between the features of the two modalities:
\begin{equation}
    m_t = ||\mathbf{f}^v_t - \mathbf{f}^a_t|| \text{,}
\end{equation}
where $\mathbf{f}^v_t$ and $\mathbf{f}^a_t$ are the features $\mathbf{F}^v$ and $\mathbf{F}^a$ at an instant $t$, respectively.

\vspace{-2mm}
\subsubsection{Attention mechanism}
To filter out irrelevant audio, such as background noise, we add an attention mechanism to focus on important pairs. We compute an attention map $\mathbf{a}$ of the same size as $\mathbf{m}$ using a cross-attention mechanism. First, we calculate an intermediate vector $\mathbf{a}'$ such that,
\begin{equation}
    a'_t = \left( \frac{\mathcal{E}^a(\mathbf{F}^a)_t \cdot \mathcal{E}^v(\mathbf{F}^v)_t}{C'} \right)  \text{,}
\end{equation}
where $\mathcal{E}^v(\cdot)$ and $\mathcal{E}^a(\cdot)$ are trainable 1D convolutions with $C'/4$ filters to reduce computation, while maintaining the dimension $T'$. We then use softmax to compute the final attention map:
\begin{equation}
    a_t = \frac{\exp(a'_t)}{\sum_{j=1}^{T'} \exp(a'_j)} \text{.}
\end{equation}
This attention map shows the degree of correlation between the audio and visual features, hence reducing the impact of unrelated noise.

Finally, the attended distance map $\hat{\mathbf m}$ is calculated as,
\begin{equation}
    \hat{\mathbf m} = \mathbf m \odot \mathbf a \text{,}
\end{equation}
where $\odot$ represents the element-wise product. It can be noted that a similar attention mechanism is used in \cite{astrid2024detecting}, but applied to the temporal rather than the spatial distance.

\subsubsection{Classifier}
The classifier $\mathcal{C}$ processes $\hat{\mathbf{m}}$ to predict the probability that the input is fake, $y$:
\begin{equation}
    y = \mathcal{C}(\hat{\mathbf m}) \text{.}
\end{equation}
The model is trained using a binary cross-entropy loss.

\vspace{-1mm}
\section{Experiments}
\label{sec:experiments}
\vspace{-1mm}
\subsection{Experimental setup}

\noindent \textbf{Dataset.} 
% Similar to~\cite{mittal2020emotions,astrid2024detecting,chugh2020not}, we train our model using a subset of the DFDC dataset~\cite{dolhansky2020deepfake}, which includes 15,300 training videos and 2,700 test videos. We balance the number of real and fake videos in the training set and keep the original class distribution in the test set. This setup is referred to as the \textit{in-dataset} protocol. 
Following prior works~\cite{mittal2020emotions,astrid2024detecting,chugh2020not}, we train on a subset of the DFDC dataset~\cite{dolhansky2020deepfake} with 15,300 training and 2,700 test videos, balancing real and fake videos in the training set while maintaining the original test set distribution (\textit{in-dataset} protocol). 
To evaluate the generalization capability of the model, we test on the FakeAVCeleb dataset~\cite{khalid2021fakeavceleb} as \textit{cross-dataset} evaluation, using 70 real and 70 fake videos, as in~\cite{khalid2021evaluation}. 
We follow the data processing of~\cite{chugh2020not}, except that we use the waveform format for audio, normalized between $-1$ to $1$. The final dimensions are $T^v=30$, $C^v=3$, $H^v=W^v=224$, $T^a=48000$.
% We preprocess the data as described in~\cite{chugh2020not}, but keep the audio in the waveform format and normalize it in the range $-1$ to $1$. The final data dimensions are the following: $T^v=30$, $C^v=3$, $H^v=W^v=224$, $T^a=48000$.

% Similar to~\cite{mittal2020emotions,astrid2024detecting,chugh2020not}, we train our dataset on partial DFDC dataset~\cite{dolhansky2020deepfake} consisting 15,300 training videos and 2,700 test videos. We sample the same number of real and fake videos for the training set and maintain the class distribution for the test set. We refer testing with DFDC test set as \textit{in-dataset} protocol. To evaluate the generalization capability of the model, we also provide \textit{cross-dataset} testing using FakeAVCeleb dataset~\cite{khalid2021fakeavceleb}. We select 70 real and 70 fake videos to build the test set following~\cite{khalid2021evaluation}. We use data preprocessing described in~\cite{chugh2020not} except we keep the audio in waveform format and min-max normalize it to range $-1$ to $1$. We end up with data of size $T^v=30$, $C^v=3$, $H^v=W^v=224$, $T^a=48000$.

\noindent \textbf{Metric.} We evaluate the performance of our method using the video-level Area Under the ROC Curve (AUC). The video-level prediction is calculated by averaging the prediction of the subsequences.

\noindent \textbf{Implementation details.} 
% We train the model for 50 epochs using the Adam optimizer \cite{kingma2014adam} with a learning rate of $10^{-3}$, weight decay of $10^{-5}$, and a mini-batch size of 8. We evaluate the model based on the lowest training loss. Unless stated otherwise, we use $T'=15$, $C'=128$, $r_{min} \approx 0$ (i.e., $l_{min} = 2$), and $r_{max} = 1$.
We train the model for 50 epochs using the Adam optimizer~\cite{kingma2014adam} with a learning rate of $10^{-3}$, weight decay of $10^{-5}$, and a batch size of $8$, evaluating based on the lowest training loss. Unless specified otherwise, we set $T'=15$, $C'=128$, $r_{min} \approx 0$ (i.e., $l_{min} = 2$), and $r_{max} = 1$.

\vspace{-2mm}
\subsection{Ablation study}
\vspace{-1mm}
In this subsection, we evaluate the importance of each component and compare different component configurations.

\vspace{-2mm}
\subsubsection{Pseudo-fake manipulation}
\vspace{-1mm}
We compare manipulation techniques from Section~\ref{subsubsec:manipulation}. As shown in Table~\ref{tab:manipulation_types}(b)-(e), all outperform the no-augmentation baseline (Table~\ref{tab:manipulation_types}(a)) in both in-dataset and cross-dataset settings. Replacing with another clip performs best, likely because it mirrors content replacement in deepfake generation. Hence, we adopt this technique for the subsequent experiments.
% We compare different manipulation techniques from Section~\ref{subsubsec:manipulation}. As shown in Table~\ref{tab:manipulation_types}(b)-(e), all techniques outperform the baseline without augmentation (Table~\ref{tab:manipulation_types}(a)) in both in-dataset and cross-dataset settings. Replacing with another clip generally performs the best, likely because it mimics the content replacement method used in many deepfake generation methods. Therefore, we use this technique in the rest of our experiments.

\begin{table}[]
\centering
\resizebox{0.9\linewidth}{!}{
\begin{tabular}{|c|l|cc|}
\hline
    & Type & AUC (In-dataset)  & AUC (Cross-dataset)   \\ \hline
(a) & No augment     & 96.42\% & 60.98\% \\
(b) & Replacing        & 96.76\% & \textbf{80.34\%} \\ 
(c) & Repeating         & 97.01\% & 67.44\% \\
(d) & Flipping           & \textbf{97.24\%} & 70.16\% \\
(e) & Translating      & 96.65\% & 62.04\% \\\hline
\end{tabular}
}
\vspace{-2mm}
\caption{Comparison in terms of AUC of different manipulation techniques used to create pseudo-fakes, as discussed in Section~\ref{subsubsec:manipulation}.}
\vspace{-2mm}
\label{tab:manipulation_types}
\end{table}

\begin{table}[]
\centering
\resizebox{0.95\linewidth}{!}{
\begin{tabular}{|c|l|cc|}
\hline
    & Temporal size & AUC (In-dataset)  & AUC (Cross-dataset)   \\ \hline
(a) & 1             & 87.20\% & 67.61\% \\
(b) & 7             & \textbf{98.03\%} & \textbf{87.02\%} \\
(c) & 15            & 96.76\% & 80.34\% \\ \hline
\end{tabular}
}
\vspace{-2mm}
\caption{Model comparison in terms of AUC with varying $T'$ values.}
\vspace{-4mm}
\label{tab:temporal_size}
\end{table}

% \subsubsection{Pseudo-fake locality}

% The finding is similar to \red{bmvc}

% \begin{table}[]
% \begin{tabular}{|c|c|cc|}
% \hline
%     & $r_{min}$ & DFDC  & FAV   \\ \hline
% (a) & $\sim$0     & 96.75 & \textbf{80.34} \\
% (b) & 0.5         & \textbf{96.82} & 73.65 \\
% (c) & 1           & 94.93 & 54.71 \\ \hline
% \end{tabular}
% \end{table}

\vspace{-2mm}
\subsubsection{Attention}
The importance of the attention mechanism in our method is demonstrated by comparing the model performance with and without it. The model incorporating attention achieves an AUC of 96.76\% in the in-dataset setting and 80.34\% in the cross-dataset setting, outperforming the model without attention, with an AUC of 96.63\% and 72.14\%, respectively. This improvement highlights the effectiveness of the attention mechanism in enhancing the model's performance.

% Table~\ref{tab:attention} shows the importance of attention mechanism in our method, where the model with attention (Table~\ref{tab:attention}(b)) outperforms the model without attention (Table~\ref{tab:attention}(a)).

% \begin{table}[]
% \centering
% \resizebox{\linewidth}{!}{
% \begin{tabular}{|c|c|cc|}
% \hline
%     & Attention & AUC (In-dataset)  & AUC (Cross-dataset)   \\ \hline
% (a) &            & 96.63\% & 72.14\% \\
% (b) & \checkmark         & \textbf{96.76\%} & \textbf{80.34\%} \\ \hline
% \end{tabular}
% }
% \caption{Ablation study evaluating the importance of attention mechanism.}
% \label{tab:attention}
% \end{table}

% \begin{table}[]
% \centering
% \begin{tabular}{|c|cc|cc|}
% \hline
%     & augmentation & attention & DFDC  & FAV   \\ \hline
% (a) &              &           & \textbf{94.62} & 63.73 \\
% (b) & \checkmark            &           & 96.63 & 72.14 \\
% (c) &              & \checkmark         & 96.42 & 60.98 \\
% (d) & \checkmark            & \checkmark         & 96.76 & \textbf{80.34} \\ \hline
% \end{tabular}
% \caption{sdafsdfsdaf}
% \end{table}

\vspace{-2mm}
\subsubsection{Distance map size}
\vspace{-1mm}

Table~\ref{tab:temporal_size} compares models with different $T'$ values. The $T'=1$ setup (Table~\ref{tab:temporal_size}(a)) performs worse than the ones set with higher $T'$ values, indicating that fine-grained distances are more effective than global distances (Table~\ref{tab:temporal_size}(b)-(c)). Interestingly, a very high $T'$, such as $T'=15$, can slightly reduce the performance compared to $T'=7$. This may be due to the model becoming too sensitive and detecting inconsistencies in low-quality real videos.

\vspace{-2mm}
\subsection{Comparison to the SOTA}
We compare our method with state-of-the-art (SOTA) audio-visual deepfake detection approaches, using the optimal setup ($T'=7$ with attention and clip replacement as an augmentation method). Our method outperforms SOTA in both in-dataset (Table~\ref{tab:in_soa}) and cross-dataset settings  (Table~\ref{tab:cross_soa}). The improved performance over FGI~\cite{astrid2024detecting} can suggest that the temporal local distance is more suitable than spatial local distance for detecting audio-visual deepfakes.

\begin{table}[]
\centering
\resizebox{0.8\linewidth}{!}{
\begin{tabular}{|l|c||l|c|}
\hline
Method          & AUC & Method    & AUC \\ \hline
MDS \cite{chugh2020not}          & 90.7\%  & VFD \cite{cheng2023voice}      & 85.1\% \\
Emotion \cite{mittal2020emotions}     & 84.4\%  & AvoiD-DF \cite{yang2023avoid}   & 94.8\% \\
BA-TFD \cite{cai2022you} & 84.6\% & SADD \cite{astrid2024statistics} & 96.7\% \\
AVT$^2$-DWF \cite{wang2024avt2}    & 89.2\%  & FGI \cite{astrid2024detecting}   & 97.7\% \\
AVFakeNet \cite{ilyas2023avfakenet}  & 86.2\%  & Ours       & \textbf{98.0\%} \\ \hline
\end{tabular}
}
\vspace{-1mm}
\caption{Comparison with SOTA in terms of AUC on DFDC (in-dataset).}
\vspace{-2mm}
\label{tab:in_soa}
\end{table}

\begin{table}[]
\centering
\resizebox{0.7\linewidth}{!}{
\begin{tabular}{|l|c||l|c|}
\hline
Method      & AUC & Method & AUC\\ \hline
MDS \cite{chugh2020not}             & 72.9\%   & SADD \cite{astrid2024statistics} & 61.4\%   \\ 
AVoiD-DF \cite{yang2023avoid}         & 82.8\%   & FGI \cite{astrid2024detecting}        & 84.5\%     \\ 
AVT$^2$-DWF \cite{wang2024avt2}       & 77.2\%     & Ours          & \textbf{87.0\%}       \\ \hline
\end{tabular}
}
\caption{Comparison with SOTA in terms of AUC  on FakeAVCeleb (cross-dataset).}
\vspace{-4mm}
\label{tab:cross_soa}
\end{table}

\section{Conclusion}
\label{sec:conclusion} 
\vspace{-2mm}
We propose detecting audio-visual deepfakes by identifying temporal inconsistencies. Our approach includes both data augmentation and architectural design strategies. For the augmentation, we experiment various manipulation techniques to create pseudo-fakes. As for the architecture, we assess the impact of local versus global distances and the role of attention mechanisms. Our method surpasses state-of-the-art approaches under both in-dataset (DFDC) and cross-dataset (FakeAVCeleb) settings.

\bibliographystyle{IEEEbib}
\bibliography{refs}

\end{document}